\documentclass[11pt,a4paper]{article}
\usepackage[hyperref]{acl2018}
\usepackage{times}
\usepackage{latexsym}
\usepackage{amsmath}
\usepackage{graphicx}
\usepackage{multirow}
\usepackage{times,latexsym}
\usepackage{url}
\usepackage{comment}
\usepackage[T1]{fontenc}

\usepackage{url}

\aclfinalcopy 

\title{Learning Cross-Lingual Sentence Representations via a \\ Multi-task Dual-Encoder Model}

\author{Muthuraman Chidambaram\thanks{\hspace{0.5em}equal contribution}, Yinfei Yang\footnotemark[1], Daniel Cer\footnotemark[1], Steve Yuan,\\
\textbf{Yun-Hsuan Sung, Brian Strope, Ray Kurzweil} \\
Google AI,  Mountain View, CA, USA \\
\texttt{\{mutty, yinfeiy, cer\}}@google.com}

\date{}

\begin{document}
\maketitle
\begin{abstract}
The scarcity of labeled training data across many languages is a significant roadblock for multilingual neural language processing. We approach the lack of in-language training data using sentence embeddings that map text written in different languages, but with similar meanings, to nearby embedding space representations. The representations are produced using a dual-encoder based model trained to maximize the representational similarity between sentence pairs drawn from parallel data. The representations are enhanced using multitask training and unsupervised monolingual corpora. The effectiveness of our multilingual sentence embeddings are assessed on a comprehensive collection of monolingual, cross-lingual, and zero-shot/few-shot learning tasks.
\end{abstract}

\section{Introduction}
\begin{minipage}{\columnwidth}
Sentence embeddings are broadly useful for a diverse collection of downstream natural language processing tasks~\citep{unec2018,infersent17,kiros2015skip,quickthought2018,subramanian18}. Sentence embeddings evaluated on downstream tasks in prior work have been trained on monolingual data, preventing them from being used for cross-lingual transfer learning. However, recent work on learning multilingual sentence embeddings has produced representations that capture semantic similarity even when sentences are written in different languages~\citep{nmt-zero-shot,Guo2018,schwenk17,singla18}. \emph{ We explore multi-task extensions of multilingual models for cross-lingual transfer learning.}
\end{minipage}

We present a novel approach for cross-lingual representation learning that combines methods for multi-task learning of monolingual sentence representations~\citep{unec2018,subramanian18} with recent work on dual encoder methods for obtaining multilingual sentence representations for bi-text retrieval~\citep{Guo2018,yang2019}. By doing so, we learn representations that maintain strong performance on the original monolingual language tasks, while \emph{simultaneously} obtaining good performance using zero-shot learning on the same task in another language. For a given language pair, we construct a multi-task training scheme using native source language tasks, native target language tasks, and a \emph{bridging translation task} to encourage sentences with identical meanings, but written in different languages, to have similar embeddings.

We evaluate the learned representations on several monolingual and cross-lingual tasks, and provide a graph-based analysis of the learned representations. Multi-task training using additional monolingual tasks is found to improve performance over models that only make use of parallel data on both cross-lingual semantic textual similarity (STS)~\citep{cer-EtAl:2017:SemEval} and cross-lingual eigen-similarity~\citep{sogard2018}. For European languages, the results show that the addition of monolingual data improves the embedding alignment of sentences and their translations. Further, we find that cross-lingual training with additional monolingual data leads to far better cross-lingual transfer learning performance.\footnote{Models based on this work are available at \url{https://tfhub.dev/} as:  universal-sentence-encoder-xling/en-de, universal-sentence-encoder-xling/en-fr, and universal-sentence-encoder-xling/en-es. A large multilingual model is available as universal-sentence-encoder-xling/many.} 

\begin{figure*}[!t]
  \centering
  \includegraphics[width=16.1cm]{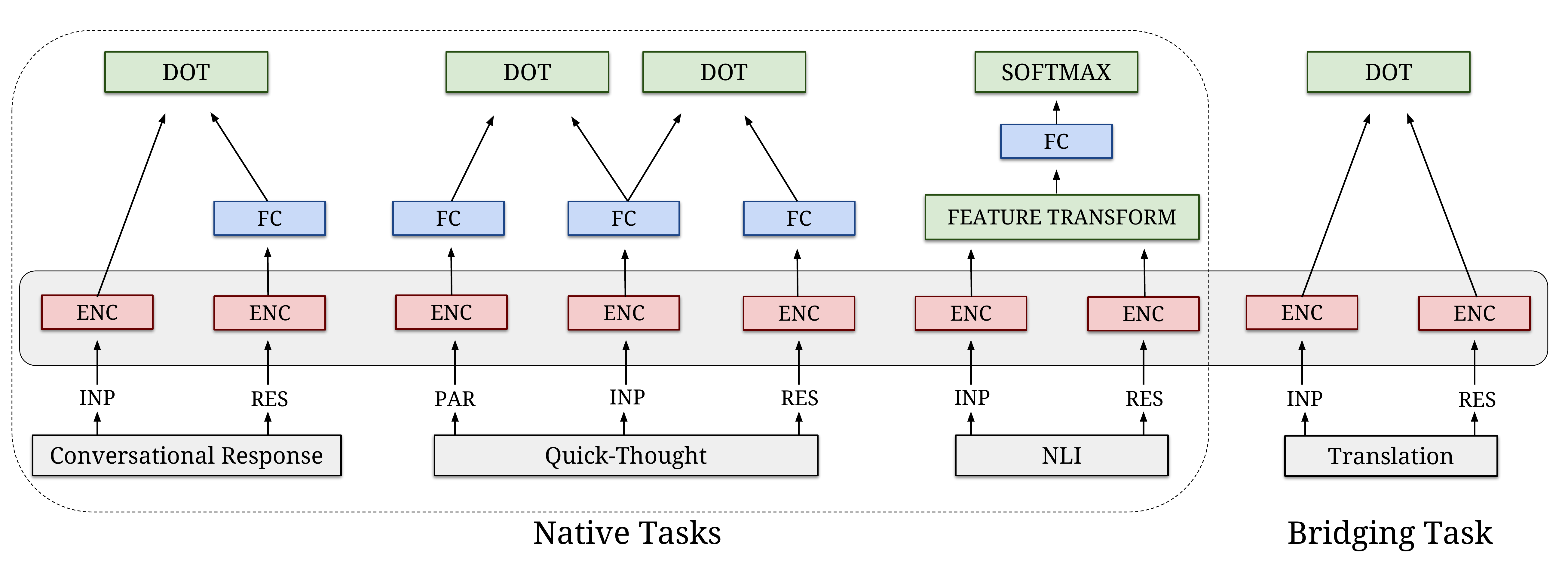}
  \caption{\label{fig:dual_encoder}
  \small
  Multi-task dual-encoder model with native tasks and a bridging translation task. The terms PAR, INP, RES refer to parent, input, and response respectively. ENC refers to the shared encoder $g$, FC refers to fully connected layers, and DOT refers to dot product. Finally, FEATURE TRANSFORM refers to the feature vector used for natural language inference.
  }
\end{figure*}

\section{Multi-Task Dual-Encoder Model}

The core of our approach is multi-task training over problems that can be modeled as ranking input-response pairs encoded via dual-encoders \cite{unec2018,henderson2017,yang2018}. Cross-lingual representations are obtained by incorporating a \emph{translation bridge task} \cite{zou-etal-2013-bilingual,gouws2015,Guo2018,yang2019}. For input-response ranking, we take an input sentence $s_i^I$ and an associated response sentence $s_i^R$, and we seek to rank $s_i^R$ over all other possible response sentences $s_j^R \in \mathcal{S}^R$. We model the conditional probability $P(s_i^R \mid s_i^I)$ as:

\begin{equation}
\begin{split}
\label{eq:prob}
    P(s_i^R \mid s_i^I) &= \frac{e^{\phi(s_i^I, s_i^R)}} {\sum_{s_j^R \in \mathcal{S}^R} e^{\phi(s_i^R, s_j^R)}} \\
    \phi(s_i^I, s_j^R) &= g^I(s_i^I)^\top g^R(s_j^R)
\end{split}
\end{equation}

Where $g^I$ and $g^R$ are the input and response sentence encoding functions that compose the dual-encoder. The normalization term in eq.\ \ref{eq:prob} is computationally intractable. We follow \citet{henderson2017} and instead choose to model an approximate conditional probability $\widetilde{P}(s_i^R \mid s_i^I)$:

\begin{equation}
\label{eq:prob2}
\widetilde{P}(s_i^R \mid s_i^I) = \frac{e^{\phi(s_i^I, s_i^R)}} {\sum_{j=1}^K e^{\phi(s_i^R, s_j^R)}}
\end{equation}

Where $K$ denotes the size of a single batch of training examples, and the $s_j^R$ corresponds to the response sentences associated with the other input sentences in the same batch as $s_i^I$. We realize $g^I$ and $g^R$ as deep neural networks that are trained to maximize the approximate log-likelihood, $\widetilde{P}(s_i^R \mid s_i^I)$, for each task.

To obtain a single sentence encoding function $g$ for use in downstream tasks, we share the first $k$ layers of the input and response encoders and treat the final output of these shared layers as $g$. The shared encoders are used with the ranking formulation above to support conversational response ranking~\cite{henderson2017}, a modified version of quick-though~\cite{quickthought2018}, and a supervised NLI task for representation learning similar to InferSent~\cite{infersent17}. To learn cross-lingual representations, we incorporate translation ranking tasks using parallel corpora for the source-target pairs: English-French (en-fr), English-Spanish (en-es), English-German (en-de), and English-Chinese (en-zh). 

The resulting model structure is illustrated in Figure \ref{fig:dual_encoder}. We note that the conversational response ranking task can be seen as a special case of Contrastive Predictive Coding (CPC)~\cite{oord2018} that only makes predictions one step into the future.

\subsection{Encoder Architecture}
\textbf{Word and Character Embeddings.} Our sentence encoder makes use of word and character $n$-gram embeddings. Word embeddings are learned end-to-end.\footnote{Using pre-trained embeddings, did not improve performance during preliminary experiments.} Character $n$-gram embeddings are learned in a similar manner and are combined at the word-level by summing their representations and then passing the resulting vector to a single feedforward layer with $tanh$ activation. We average the word and character embeddings before providing them as input to $g$.

\textbf{Transformer Encoder.} The architecture of the shared encoder $g$ consists of three stacked transformer sub-networks,\footnote{We tried up to six stacked transformers, but did not notice a significant difference beyond three.} each containing the feed-forward and multi-head attention sub-layers described in \citet{transformer2017}. The transformer output is a variable-length sequence.  We average encodings of all sequence positions in the final layer to obtain our sentence embeddings. This embedding is then fed into different sets of feedforward layers that are used for each task. For our transformer layers, we use 8 attentions heads, a hidden size of 512, and a filter size of 2048.

\subsection{Multi-task Training Setup}

We employ four unique task types for each language pair in order to learn a function $g$ that is capable of strong cross-lingual semantic matching and transfer learning performance for a source-target language pair while also maintaining monolingual task transfer performance. Specifically, we employ: \textit{(i) conversational response prediction, (ii) quick thought, (iii) a natural language inference}, and \textit{(iv) translation ranking} as the bridge task. For models trained on a single language pair (e.g., en-fr), six total tasks are used in training, as the first two tasks are mirrored across languages.\footnote{We note that our architecture can scale to models trained on $> 2$ languages. Preliminary experiments using more than two languages achieve promising results, but we consider fully evaluating models trained on larger collections of languages to be outside the scope of the current work.}

\textbf{Conversational Response Prediction.} We model the conversational response prediction task in the same manner as \citet{yang2018}. We minimize the negative log-likelihood of $\widetilde{P}(s_i^R \mid s_i^I)$, where $s_i^I$ is a single comment and $s_i^R$ is its associated response comment. For the response side, we model $g^R(s_i^R)$ as $g(s_i^R)$ followed by two fully-connected feedforward layers of size 320 and 512 with $tanh$ activation. For the input representation, however, we simply let $g^I(s_i^I) = g(s_i^I)$.\footnote{In early experiments, letting the optimization of the conversational response task more directly influence the parameters of the underlying sentence encoder $g$ led to better downstream task performance.}

\textbf{Quick Thought.} We use a modified version of the Quick Thought task detailed by \citet{quickthought2018}. We minimize the sum of the negative log-likelihoods of $\widetilde{P}(s_i^R \mid s_i^I)$ and $\widetilde{P}(s_i^P \mid s_i^I)$, where $s_i^I$ is a sentence taken from an article and $s_i^P$ and $s_i^R$ are its predecessor and successor sentences, respectively. For this task, we model all three of $g^P(s_i^P)$, $g^I(s_i^I)$, and $g^R(s_i^R)$ by $g$ followed by separate, fully-connected feedforward layers of size 320 and 512 and using $tanh$ activation.

\textbf{Natural Language Inference (NLI).} We also include an \emph{English-only} natural language inference task~\cite{bowman2015}. For this task, we first encode an input sentence $s_i^I$ and its corresponding response hypothesis $s_i^R$ into vectors $u_1$ and $u_2$ using $g$. Following \newcite{infersent17}, the vectors $u_1$, $u_2$ are then used to construct a relation feature vector $(u_1, u_2, |u_1-u_2|, u_1*u_2)$, where $(\cdot)$ represents concatenation and $*$ represents element-wise multiplication. The relation vector is then fed into a single feedforward layer of size 512 followed by a softmax output layer that is used to perform the 3-way NLI classification.

\textbf{Translation Ranking.} Our translation task setup is identical to the one used by \citet{Guo2018} for bi-text retrieval. We minimize the negative log-likelihood of $\widetilde{P}(s_i \mid t_i)$, where ($s_i$, $t_i$) is a source-target translation pair. Since the translation task is intended to align the sentence representations of the source and target languages, we do not use any kind of task-specific feedforward layers and instead use $g$ as both $g^I$ and $g^R$. Following \citet{Guo2018}, we append 5 incorrect translations that are semantically similar to the correct translation for each training example as ``hard-negatives''. Similarity is determined via a version of our model trained only on the translation ranking task. We did not see additional gains from using more than 5 hard-negatives.

\section{Experiments}

\begin{table*}[ht!]
\fontsize{10.3}{11.5}\selectfont
\begin{center}
    \begin{tabular}{|c|c|c|c|c|c|c|c|}
    \hline
    \multirow{2}{*}{\bf Model} &  \multirow{2}{*}{\bf MR} & \multirow{2}{*}{\bf CR} & \multirow{2}{*}{\bf SUBJ} & \multirow{2}{*}{\bf MPQA} & \multirow{2}{*}{\bf TREC} & \multirow{2}{*}{\bf SST} & {\bf STS Bench}\\
    & & & & & & & {\bf (dev / test)} \\
    \hline
    \multicolumn{8}{|c|}{\emph{Cross-lingual Multi-task Models}} \\
    \hline
    en-fr & 77.9  &  82.9  &  95.5  &  \underline{89.3}  &  95.3  &  84.0 & 0.803 / 0.763 \\ 
    en-es & \underline{80.1}  &  \underline{85.9}  &  94.6  &  86.5  &  96.2  &  \underline{85.2} & \underline{0.809} / \underline{0.770} \\ 
    en-de & 78.8  &  84.0  &  \textbf{95.9}  &  87.6  &  96.1  &  85.0 & 0.802 / 0.764 \\ 
    en-zh & 76.1  &  83.4  &  93.0  &  86.4  &  \textbf{97.7}  &  81.4 & 0.791 / \underline{0.770} \\
    \hline
    \multicolumn{8}{|c|}{\emph{Translation-ranking Models}} \\
    \hline
    en-fr &  68.7  &  79.3  &  87.0  &  81.8  &  89.4  &  74.2 & 0.668 / 0.558 \\ 
    en-es &  67.7  &  75.7  &  83.5  &  86.0  &  94.4  &  72.6 & 0.669 / 0.631 \\ 
    en-de &  67.8  &  75.2  &  84.4  &  83.6  &  86.8  &  74.6 & 0.673 / 0.632 \\ 
    en-zh &  73.6  &  78.5  &  88.1  &  88.2  &  96.1  &  77.1 & 0.779 / 0.761 \\
    \hline
    \multicolumn{8}{|c|}{\emph{Prior Work}} \\
    \hline
    CPC~\cite{oord2018} & 76.9  & 80.1  & 91.2 & 87.7  & \underline{96.8} & -- & --     \\
    USE Trans.~\cite{unec2018} & \textbf{81.4} & \textbf{87.4} & 93.9 & 87.0 & 92.5 & 85.4 & \textbf{0.814} / \textbf{0.782} \\
    QT~\cite{quickthought2018} & 82.4 & 86.0 & \underline{94.8} & \textbf{90.2} & 92.4 & \textbf{87.6} & -- \\
    InferSent~\cite{infersent17} & 81.1 & 86.3 & 92.4 & \textbf{90.2} & 88.2 & 84.6 & 0.801 / 0.758 \\
    ST\ LN~\cite{kiros2015skip} & 79.4 & 83.1 & 93.7 & 89.3 & -- & -- & -- \\
    \hline
    \end{tabular}
\end{center}
\caption{Performance on classification transfer tasks from SentEval \citep{conneau2018senteval}.}
\label{tab:trans-model-performance}
\end{table*}

\subsection{Corpora}
Training data is composed of Reddit, Wikipedia, Stanford Natural Language Inference (SNLI), and web mined translation pairs. For each of our datasets, we use 90\% of the data for training, and the remaining 10\% for development/validation.

\subsection{Model Configuration}
In all of our experiments, multi-task training is performed by cycling through the different tasks (translation pairs, Reddit, Wikipedia, NLI) and performing an optimization step for a single task at a time.
We train all of our models with a batch size of 100 using stochastic gradient descent with a learning rate of 0.008.
All of our models are trained for 30 million steps.
All input text is tree-bank style tokenized prior to being used for training.
We build a vocab containing 200 thousand unigram tokens with 10 thousand hash buckets for out-of-vocabulary tokens.
The character $n$-gram vocab contains 200 thousand hash buckets used for 3 and 4 grams.
Both the word and character $n$-gram embedding sizes are 320.
All hyperparameters are tuned based on the development portion (random 10\% slice) of our training sets.
As an additional training heuristic, we multiply the gradient updates to the word and character embeddings by a factor of 100.\footnote{We tried different orders of magnitude for the multiplier and found 100 to work the best.} We found that using this embedding gradient multiplier alleviates vanishing gradients and greatly improves training.

We compare the proposed cross-lingual multi-task models, subsequently referred to simply as ``multi-task", with baseline models that are trained using only the translation ranking task, referred to as ``translation-ranking'' models.

\subsection{Model Performance on English Downstream Tasks}
We first evaluate all of our cross-lingual models on several downstream English tasks taken from SentEval \citep{conneau2018senteval} to verify the impact of cross-lingual training. Evaluations are performed by training single hidden-layer feedforward networks on top of the 512-dimensional embeddings taken from the frozen models.
Results on the tasks are summarized in Table \ref{tab:trans-model-performance}. We note that cross-lingual training does not hinder the effectiveness of our encoder on English tasks, as the multi-task models are close to state-of-the-art in each of the downstream tasks. For the Text REtrieval Conference (TREC) eval, we actually find that our multi-task models outperform the previous state-of-the-art by a sizable amount.

We observe the en-zh translation-ranking models perform significantly better on the downstream tasks than the European language pair translation-ranking models. The en-zh models are possibly less capable of exploiting grammatical and other superficial similarities and are forced to rely on semantic representations. Exploring this further may present a promising direction for future research.

\begin{table*}[ht!]
\fontsize{11}{13.1}\selectfont
\begin{center}
    \begin{tabular}{|c|c|c|c|c|c|}
    \hline
    
    \multirow{2}{*}{\bf Model}   &  \multicolumn{5}{c|}{\bf STS Benchmark (dev / test)} \\ \cline{2-6}
                                 & {\bf en} & {\bf fr} & {\bf es} & {\bf de} & \bf{zh} \\ \hline
    Multi-task en-fr             & 0.803 / 0.763 & \textbf{0.777} / \textbf{0.738} &   --  &   --  &   --   \\
    Trans.-ranking en-fr         & 0.668 / 0.558 & 0.641 / 0.579 &   --  &   --  &   --   \\ \hline
    Multi-task en-es             & \textbf{0.809} / \textbf{0.770} &   --  & \textbf{0.779} / \textbf{0.744} &   --  &   --   \\
    Trans.-ranking en-es         & 0.669 / 0.631 &   --  & 0.622 / 0.611 &   --  &   --   \\ \hline
    Multi-task en-de             & 0.802 / 0.764 &   --  &   --  & \textbf{0.768} / \textbf{0.722} &   --   \\
    Trans.-ranking en-de         & 0.673 / 0.632 &   --  &   --  & 0.630 / 0.526 &   --   \\ \hline
    Multi-task en-zh             & 0.791 / \textbf{0.770} &   --  &   --  &   --  & 0.730 / \textbf{0.705}  \\
    Trans.-ranking en-zh         & 0.779 / 0.761 &   --  &   --  &   --  & \textbf{0.733} / 0.701  \\ \hline
    \end{tabular}
\end{center}
\caption{Pearson's correlation coefficients on STS Benchmark (dev / test). The first column shows the results on the original STS Benchmark data in English, French, and Spanish. }
\label{tab:multilingual-sts}
\end{table*}

\subsection{Cross-lingual Retrieval}
We evaluate both the multi-task and translation-ranking models' efficacy in performing cross-lingual retrieval by using held-out translation pair data.
Following \citet{Guo2018} and \citet{henderson2017}, we use precision at N (P@N) as our evaluation metric. Performance is scored by checking if a source sentence's target translation ranks\footnote{Translation ranking scores are obtained by the dot product of source and target representations} in the top $N$ scored candidates when considering $K$ other randomly selected target sentences. We set $K$ to 999.
Similar to \citet{Guo2018}, we observe using a small value of $K$, such as $K=99$ from \citet{henderson2017}, results in all metrics quickly obtaining $>99\%$ P@1.\footnote{999 is smaller than the 10+ million used by \citet{Guo2018}, but it allows for good discrimination between models without requiring a heavier and slower evaluation framework}

The translation-ranking model is a strong baseline for identifying correct translations, with 95.4\%, 87.5\%, 97.5\%, and 99.7\% P@1 for en-fr, en-es, en-de, and en-zh retrieval tasks, respectively.
The multi-task model performs almost identical with 95.1\%, 88.8\%, 97.8\%, and 99.7\% P@1, which provides empirical justification that it is possible to maintain cross-lingual embedding space alignment despite training on additional monolingual tasks for each individual language.\footnote{We also experimented with  P@3 and P@10, the results are identical.} Both model types surprisingly achieve particularly strong ranking performance on en-zh. Similar to the task transfer experiments, this may be due to the en-zh models having an implicit inductive bias to rely more heavily on semantics rather than more superficial aspects of sentence pair similarity.

\subsection{Multilingual STS}

Cross-lingual representations are evaluated on semantic textual similarity (STS) in French, Spanish, German, and Chinese. To evaluate Spanish-Spanish (es-es) STS, we use data from track 3 of the SemEval-2017 STS shared task~\cite{cer-EtAl:2017:SemEval}, containing 250 Spanish sentence pairs.
We evaluate English-Spanish (en-es) STS using STS 2017 track 4(a),\footnote{The en-es task is split into track 4(a) and track 4(b). We only use track 4(a) here. Track 4(b) contains sentence pairs from WMT with only one annotator for each pair. Previously reported numbers are particularly low for track 4(b), which may suggest either distributional or annotation differences between this track and other STS datasets.} which contains 250 English-Spanish sentence pairs.

Beyond English and Spanish, however, there are no standard STS datasets available for the other languages explored in this work. 
As such, we perform an additional evaluation on a translated version of the STS Benchmark~\cite{cer-EtAl:2017:SemEval} for French, Spanish, German, and Chinese.
We use Google's translation system to translate the STS Benchmark sentences into each of these languages.
We believe the results on the translated STS Benchmark evaluation sets are a reasonable indicator of multilingual semantic similarly performance, particularly since the NMT encoder-decoder architecture for translation differs significantly from our dual-encoder approach.

Following \citet{unec2018}, we first compute the sentence embeddings $u,v$ for an STS sentence pair, and then score the sentence pair similarity based on the angular distance between the two embedding vectors, $-\arccos\Big(\frac{u v}{||u|| ~ ||v||}\Big)$. 
Table \ref{tab:multilingual-sts} shows Pearson's $r$ on the STS Benchmark for all models.
The first column shows the trained model performance on the original English STS Benchmark.
Columns 2 to 5 provide the performance on the remaining languages.
Multi-task models perform better than the translation ranking models on our multilingual STS Benchmark evaluation sets.
Table \ref{tab:sts17} provides the results from the en-es models on the SemEval-2017 STS *-es tracks.
The multi-task models achieve 0.827 Pearson's $r$ for the es-es task and 0.769 for the en-es task.
As a point of reference, we also list the two best performing STS systems, ECNU~\cite{ecnu} and BIT~\cite{bit}, as reported in \citet{cer-EtAl:2017:SemEval}.
Our results are very close to these state-of-the-art feature engineered and mixed systems.

\begin{table}
\begin{center}
    \begin{tabular}{|c|c|c|}
    \hline
    \multirow{2}{*}{\bf Model}   &  \multicolumn{2}{c|}{\bf STS (SemEval 2017)} \\ \cline{2-3}
               & {\bf es-es} & {\bf en-es} \\ \hline
    Multi-task & \underline{0.827} & \underline{0.769} \\
    Trans.-ranking & 0.642 & 0.587 \\ \hline
    ECNU & \textbf{0.856} & \textbf{0.813} \\
    BIT & 0.846 & 0.749 \\ \hline
    \end{tabular}
\end{center}
\caption{Pearson's $r$ on track 3 (es-es) and track 4(a) (en-es) of the SemEval-2017 STS shared task.}
\label{tab:sts17}
\end{table}

\begin{table*}[ht!]
\fontsize{10.3}{12}\selectfont
\begin{center}
    \begin{tabular}{|c|c|c|c||c|c|c|c|c|}
    \hline
    \multirow{2}{*}{\bf Model} & \multicolumn{3}{c||}{\bf SNLI-X} &  \multicolumn{5}{c|}{\bf XNLI}\\ \cline{2-9}
    & {\bf en} & {\bf fr} & {\bf es} & {\bf en} & {\bf fr} & {\bf es} & {\bf de} & {\bf zh}\\ \hline
    Multi-task en-fr                  & \underline{84.2} & \textbf{74.0} &  --  & \textbf{71.6} & \textbf{64.4} &  --  & --   & --   \\
    Multi-task en-es                  & 83.9 & --   & 75.9 & 70.2 &  --  & \textbf{65.2} & --   & --   \\
    Multi-task en-de                  & 84.1 & --   &  --  & 71.5 &  --  &  --  & \textbf{65.0} & --   \\
    Multi-task en-zh                  & 83.7 & --   &  --  & 69.2 &  --  &  --  & --   & \textbf{62.8} \\ \hline
    NMT en-fr~\cite{nmt-zero-shot}   & \textbf{84.4} & 73.9 &  --  &      &  --  &  --  & --   & --   \\
    XNLI-CBOW zero-shot~\cite{xnli}               &  --  & --   &  --  & 64.5 & 60.3 & 60.7 & 61.0 & 58.8 \\ \hline \hline
    \multicolumn{9}{|c|}{\emph{Non zero-shot baselines}} \\ \hline
    XNLI-BiLSTM-last~\cite{xnli} &  --  &  --  &  --  & 71.0 & 65.2 & 67.8 & 66.6 & 63.7 \\
    XNLI-BiLSTM-max~\cite{xnli} &  --  &  --  &  --  & \textbf{73.7} & \textbf{67.7} & \textbf{68.7} & \textbf{67.7} & \textbf{65.8} \\
    \hline
    \end{tabular}
\end{center}
\caption{Zero-shot classification accuracy (\%) on SNLI-X and XNLI datasets. Cross-lingual transfer models are training on English only NLI data and then evaluated on French (fr), Spanish (es), German (de) and Chinese (zh) evaluation sets.}
\label{tab:zero-sxnli}
\end{table*}

\section{Zero-shot Classification}

To evaluate the cross-lingual transfer learning capabilities of our models, we examine performance of the multi-task and translation-ranking encoders on zero-shot and few-shot classification tasks.

\subsection{Multilingual NLI}
We evaluate the zero-shot classification performance of our multi-task models on two multilingual natural language inference (NLI) tasks. However, prior to doing so, we first train a modified version\footnote{Training with additional MultiNLI data did not significantly impact SNLI or downstream task performance.} of our multi-task models that also includes training on the English Multi-genre NLI (MultiNLI) dataset of \citet{multinli} in addition to SNLI. We train with MultiNLI to be consistent with the baselines from prior work.

We make use of the professionally translated French and Spanish SNLI subset created by \citet{xling-snli} for an initial cross-lingual zero-shot evaluation of French and Spanish. We refer to these translated subsets as SNLI-X.
There are 1,000 examples in the subset for each language.
To evaluate, we feed the French and Spanish examples into the pre-trained English NLI sub-network of our multi-task models.

We additionally make use of the XNLI dataset of \citet{xnli}, which provides multilingual NLI evaluations for Spanish, French, German, Chinese and more. There are 5,000 examples in each XNLI test set, and zero-shot evaluation is once again done by feeding non-English examples into the pre-trained English NLI sub-network.

Table \ref{tab:zero-sxnli} lists the accuracy on the English SNLI test set as well as on SNLI-X and XNLI for all of our multi-task models.
The original English SNLI accuracies are around 84\% for all of our multi-task models, indicating that English SNLI performance remains stable in the multi-task training setting.
The zero-shot accuracy on SNLI-X is around 74\% for both the en-fr and en-es models. The zero-shot accuracy on XNLI is around 65\% for en-es, en-fr, and en-de, and around 63\% for en-zh, thereby significantly outperforming the pretrained sentence encoding baselines (X-CBOW) described in \citet{xnli}. The X-CBOW baselines use fixed sentence encoders that are the result of averaging tuned multilingual word embeddings. 

Row 4 of Table \ref{tab:zero-sxnli} shows the zero-shot French NLI performance of \citet{nmt-zero-shot}, which is a state-of-the-art zero-shot NLI classifier based on multilingual NMT embeddings.
Our multi-task model shows comparable performance to the NMT-based model in both English and French.

\subsection{Amazon Reviews}

\begin{table}[t]

\begin{center}
\resizebox{\columnwidth}{!}{%
    \begin{tabular}{|c|c|c|c|}
    \hline
    {\bf Model} & {\bf en} & {\bf fr} & {\bf de} \\ \hline
    Multi-task en-fr          & \textbf{87.4} & \textbf{82.3} &   --  \\ 
    Translation-ranking en-fr    & 74.4 & 66.3 &   --  \\ \hline
    Multi-task en-de          & \textbf{87.1} &  --  & \textbf{81.0} \\
    Translation-ranking en-de    & 73.8 & -- & 67.0 \\ \hline
    \citet{nmt-zero-shot}  (NMT en-fr)   & 83.2 & 81.3 &   --  \\
    \hline
    \end{tabular}
}
\end{center}
\caption{Zero-shot sentiment classification accuracy(\%) on non-English Amazon review test data after training on English only Amazon reviews.}
\label{tab:amazon-zero-shot}
\vspace{-0.05cm}
\end{table}

\begin{table*}[t]

\begin{center}
    \begin{tabular}{|c|l r|c|c|c|c|c|c|}
    \hline
    {\bf Target} &  \multicolumn{2}{|c|}{\multirow{2}{*}{\bf Model}} & \multicolumn{6}{c|}{\bf \% available fr/de data} \\
    \cline{4-9}
    \hspace{0.5em}{\bf Language}\hspace{0.5em} & & & 0\% &10\% & 20\% & 40\% & 80\% & 100\% \\ \hline
    \multirow{2}{*}{French} & 
    \hspace{0.25em}Few-shot & 100\% en + X\% fr& 82.3 & \textbf{84.4} & \textbf{84.4} & \textbf{84.8} & \textbf{85.2} & \textbf{85.8}  \\ 
    & \hspace{0.25em}Monolingual & 0\% en + X\% fr &   -- & 79.2 & 80.0 & 82.7 & 84.3 & 84.9 \\ \hline
    
    \multirow{2}{*}{German} &
    \hspace{0.25em}Few-shot & 100\% en + X\% de & 81.0 & \textbf{81.6} & \textbf{83.3} & \textbf{84.0} & \textbf{84.7} & \textbf{84.5} \\ 
    & \hspace{0.25em}Monolingual & 0\% en + X\% de &  --  & 75.5 & 77.7 & 81.6 & 83.5 & 84.4 \\
    \hline
    \end{tabular}
\end{center}
\caption{Sentiment classification accuracy(\%) on target language Amazon review test data after training on English Amazon review data and a portion of French of German data. The second row shows the percent of French (fr) or German (de) data is used for training in each model.}
\label{tab:amazon-few-shot}
\end{table*}

\textbf{Zero-shot Learning.} We also conduct a zero-shot evaluation based on the Amazon review data extracted by \citet{stein:2010k}.  Following \citet{stein:2010k}, we preprocess the Amazon reviews and convert the data into a binary sentiment classification task by considering reviews with strictly more than three stars as positive and  less than three stars as negative. 
Reviews contain a summary field and a text field, which we concatenate to produce a single input.
Since our models are trained with sentence lengths clipped to 64, we only take the first 64 tokens from the concatenated text as the input.
There are 6,000 training reviews in English, which we split into 90\% for training and 10\% for development.

We first encode inputs using the pre-trained multi-task and translation-ranking encoders and feed the encoded vectors into a 2-layer feed-forward network culminating in a softmax layer. We use hidden layers of size 512 with $tanh$ activation functions.
We use Adam for optimization with an initial learning rate of 0.0005 and a learning rate decay of 0.9 at every epoch during training. We use a batch size of 16 and train for 20 total epochs in all experiments. We freeze the cross-lingual encoder during training.
The model architecture and parameters are tuned on the development set.

We first train the classifier on English data, and then evaluate it on the 6,000 French and German Amazon review test examples.
The results are summarized in Table \ref{tab:amazon-zero-shot}.
On the English test set, accuracy of the en-fr model is 87.4\% with the en-de model achieving 87.1\%. Both models achieve zero-shot accuracy on their respective non-English datasets that is above 80\%.
The translation-ranking models again perform worse on all metrics.
Once again we compare the proposed model with \citet{nmt-zero-shot}, and find that our zero-shot performance has a reasonable gain on the French test set.\footnote{\citet{nmt-zero-shot} also train a shallow classifier, but use only review text and truncate their inputs to 200 tokens. Our setup is slightly different, as our models can take a maximum of only 64 tokens.}

\textbf{Few-shot Learning.} We further evaluate the proposed multi-task models via few-shot learning, by training on English reviews and only a portion of French and German reviews.
Our few-shot models are compared with baselines trained on French and German reviews only.
Table \ref{tab:amazon-few-shot} provides the classification accuracy of the few-shot models, where the second row indicates the percent of French and German data that is used when training each model.
With as little as 20\% of the French or German training data, the few-shot models perform nearly as well compare to the baseline models trained on 100\% of the French and German data.
Adding more French and German training data leads to further improvements in few-shot model performance, with the few-shot models reaching 85.8\% accuracy in French and 84.5\% accuracy in German, when using all of the French and German data. The French model notably performs $+0.9\%$ better when being trained on a combination of the English and French reviews rather than on the French reviews alone.

\section{Analysis of Cross-lingual Embedding Spaces}
Motivated by the recent work of \citet{sogard2018} studying the graph structure of multilingual word representations, we perform a similar analysis for our learned cross-lingual sentence representations. We take $N$ samples of size $K$ from the language pair translation data and then encode these samples using the corresponding multi-task and translation-ranking models. We then compute pairwise distance matrices within each sampled set of encodings, and use these distance matrices to construct graph Laplacians.\footnote{See \citet{graphlap} for an overview of graph Laplacians.} We obtain the similarity $\Psi(S, T)$ between each model's source and target language embedding by comparing the eigenvalues of the source language graph Laplacians to the eigenvalues of the target language graph Laplacians:

\begin{equation}
\label{eq:laplacian}
\Psi(S, T) = \frac{1}{N} \sum_{i=1}^N \sum_{j=1}^K (\lambda_j(L_i^{(s)}) - \lambda_j(L_i^{(t)}))^2
\end{equation}

Where $L_i^{(s)}$ and $L_i^{(t)}$ refer to the graph Laplacians of the source language and target language sentences obtained from the $i^{th}$ sample of source-target translation pairs. A smaller value of $\Psi(S, T)$ indicates higher eigen-similarity between the source language and target language embedding subsets. Following \citet{sogard2018} we use a sample size of $K = 10$ translation pairs, but we choose to draw $N = 1,000$ samples instead of $N = 10$, as was done in \citet{sogard2018}. We found $\Psi(S, T)$ has very high variance at $N = 10$. The computed values of $\Psi(S, T)$ for our multi-task and translation-ranking models are summarized in Table \ref{tab:frobsim}.

We find that the source and target embedding subsets constructed from the multi-task models exhibit greater average eigen-similarity than those resulting from the translation-ranking models for the European source-target language pairs, and observe the opposite for the English-Chinese models (en-zh). As a curious discrepancy, we believe further experiments looking at eigen-similarity across languages could yield interesting results and language groupings.

Eigen-similarity trends with better performance for the European language pair multi-task models on the cross-lingual transfer tasks. A potential direction for future work could be to introduce regularization penalties based on graph similarity during multi-task training. Interestingly, we also observe that the eigen-similarity gaps between the multi-task and translation-ranking models are not uniform across language pairs. Thus, another direction could be to further study differences in the difficulty of aligning different source-target language embeddings.

\subsection{Discussion on Input Representations}
Our early explorations using a combination of character $n$-gram embeddings and word embeddings vs.\ word embeddings alone as the model input representation suggest using word-embeddings only performs just slightly worse (one to two absolute percentage points) on the dev sets for the training tasks. The notable exception is the word-embedding only English-German models tend to perform much worse on the dev sets for the training tasks involving German. This is likely due to the prevalence of compound words in German and represents an interesting difference for future exploration.

\begin{table}[t]
\begin{center}
\resizebox{\columnwidth}{!}{%
    \begin{tabular}{|c|c|c|c|c|}
    \hline
    {\bf Model} & {\bf  en-fr} & {\bf en-es} & {\bf en-de} & {\bf en-zh}\\ \hline
    multi-task          & \textbf{0.592} & \textbf{0.526} & \textbf{0.761}  & 2.366 \\ 
    trans.-ranking      & 1.036 & 0.572 & 2.187 & \textbf{0.393} \\
    \hline
    \end{tabular}
}
\end{center}
\caption{Average eigen-similarity values of source and target embedding subsets.}
\label{tab:frobsim}
\end{table}

We subsequently explored training versions of our cross-lingual models using a SentencePiece vocabulary~\cite{sentencepiece}, a set of largely sub-word tokens (characters and word chunks) that provide good coverage of an input dataset. Multilingual models for a single language pair (e.g., en-de) trained with SentencePiece performed similarly on the training dev sets to the models using character $n$-grams. However, when more languages are included in a single model (e.g., a single model that covers en, fr, de, es, and zh), SentencePiece tends to perform worse than using a combination of word and character $n$-gram embeddings. Within a larger joint model, SentencePiece is particularly problematic for languages like zh, which end up getting largely tokenized into individual characters.

\section{Conclusion}

Cross-lingual multi-task dual-encoder models are found to learn representations that achieve strong within language and cross-lingual transfer learning performance. By training English-French, English-Spanish, English-German, and English-Chinese multi-task models, we achieve near-state-of-the-art or state-of-the-art performance on a variety of English tasks, while also being able to produce similar caliber results in zero-shot cross-lingual transfer learning tasks. Further, cross-lingual multi-task training is shown to improve performance on some downstream English tasks (TREC). We believe that there are many possibilities for future explorations of cross-lingual model training and that such models will be foundational as language processing systems are tasked with increasing amounts of multilingual data.

\section*{Acknowledgments}

We thank the anonymous reviewers and our teammates from Descartes and other Google groups for their feedback and suggestions.

\bibliography{acl2018}
\bibliographystyle{acl_natbib}

\end{document}